\newcommand{\subfigimg}[3][,]{%
  \setbox1=\hbox{\includegraphics[#1]{#3}}
  \leavevmode\rlap{\usebox1}
  \rlap{\hspace*{18pt}\raisebox{\dimexpr\ht1-1.27\baselineskip}{#2}}
  \phantom{\usebox1}
}

\documentclass{article}
\usepackage{graphicx} 
\usepackage{multirow}
\usepackage{subcaption,amsmath}

\usepackage[preprint]{neurips_2025}

\usepackage[utf8]{inputenc} 
\usepackage[T1]{fontenc}    
\usepackage{hyperref}       
\usepackage{url}            
\usepackage{booktabs}       
\usepackage{amsfonts}       
\usepackage{nicefrac}       
\usepackage{microtype}      
\usepackage{xcolor}         

\title{Foundation model for mass spectrometry proteomics}

\author{%
  Justin Sanders $^{1}$\thanks{Equal contributions} \\
  \texttt{jsander1@cs.washington.edu}
  \And
  Melih Yilmaz $^{1*}$ \\
  \texttt{melih@cs.washington.edu}          
  \AND
  Jacob H. Russell $^{2}$ \\
  \texttt{rus22@uw.edu}
  \And
  Wout Bittremieux $^{3}$ \\
  \texttt{wout.bittremieux@uantwerpen.be}
  \And
  William E. Fondrie $^{4}$ \\
  \texttt{wfondrie@talus.bio}
  \AND
  Nicholas M. Riley $^{2}$\\
  \texttt{nmriley@uw.edu}
  \And
  Sewoong Oh $^{1}$\\
  \texttt{sewoong@cs.washington.edu}
  \And
  William Stafford Noble $^{1,5}$\\
  \texttt{wnoble@uw.edu}
\And
\\
$^1$ Paul G.\ Allen School of Computer Science and Engineering, University of Washington \\ \quad $^2$ Department of Chemistry, University of Washington \\ \quad $^3$ Adrem Data Lab, University of Antwerp \quad $^4$ Talus Bioscience \\ \quad $^5$ Department of Genome Sciences, University of Washington
}

\begin{document}

\maketitle

\begin{abstract}
Mass spectrometry is the dominant technology in the field of proteomics, enabling high-throughput analysis of the protein content of complex biological samples. Due to the complexity of the instrumentation and resulting data, sophisticated computational methods are required for the processing and interpretation of acquired mass spectra. Machine learning has shown great promise to improve the analysis of mass spectrometry data, with numerous purpose-built methods for improving specific steps in the data acquisition and analysis pipeline reaching widespread adoption. Here, we propose unifying various spectrum prediction tasks under a single foundation model for mass spectra. To this end, we pre-train a spectrum encoder using \textit{de novo} sequencing as a pre-training task. We then show that using these pre-trained spectrum representations improves our performance on the four downstream tasks of spectrum quality prediction, chimericity prediction, phosphorylation prediction, and glycosylation status prediction. Finally, we perform multi-task fine-tuning and find that this approach improves the performance on each task individually. Overall, our work demonstrates that a foundation model for tandem mass spectrometry proteomics trained on \textit{de novo} sequencing learns generalizable representations of spectra, improves performance on downstream tasks where training data is limited, and can ultimately enhance data acquisition and analysis in proteomics experiments. 
\end{abstract}

\section{Introduction}

In recent years, foundation models have emerged as a powerful machine learning paradigm for various problem domains \cite{radford2018improving,radford:learning,bommasani2021opportunities}.
These models are trained to learn rich latent representations of input modalities from large datasets of unlabeled or weakly labeled data (e.g., online text, protein sequences in public repositories) using pre-training tasks such as masked language modeling. 
The trained model can subsequently be used to perform a variety of downstream tasks, relying on the same input modality with little or no supervised fine-tuning for the specific task in question.
In many cases, a foundation model will outperform its peers trained only with supervision and will achieve better performance with a relatively small amount of supervised fine-tuning.

Motivated by the success of foundation models in language, vision, and multi-modal tasks, we develop a foundation model for tandem mass spectrometry proteomics.
Typically, the primary analysis problem for mass spectrometry data is the spectrum annotation problem, i.e., assigning to each spectrum the peptide sequence which generated it.
However, there are a variety of other important analysis questions including $(i)$ predicting whether a given spectrum is generated by an identifiable peptide as opposed to noise or contamination, $(ii)$ detecting whether a spectrum contains signal from a mixture of multiple analytes at once, and $(iii)$ predicting whether a spectrum contains signal for a peptide carrying a particular post-translational modification (PTM, i.e., a specific molecular group attached to one of the amino acids in the peptide). 
For many tasks, insufficient training data and noisy training labels make it challenging to learn a rich understanding of mass spectra in isolation for each task.
We hypothesized that learned spectrum embeddings, pre-trained on a large dataset of high-confidence spectrum annotations, may prove a valuable starting point for each of these different downstream tasks.
We empirically demonstrate that our pretrained spectrum encoder, Casanovo Foundation, improves  downstream task performance over task-specific state-of-the-art models on a variety of prediction tasks.

\section{Background and related work}

Currently, tandem mass spectrometry is the only high-throughput method for systematically analyzing the full protein content of biological samples \cite{maccoss2023sampling}, driving breakthroughs in disease biomarker discovery, drug development, and analysis of PTMs \cite{aebersold:mass-spectrometric}. 
In a standard tandem mass spectrometry experiment, proteins are digested into short peptides, ionized, and fragmented. The mass-to-charge ratios (\textit{m/z}) of the resulting fragment ions are then measured very precisely by the instrument. 
This process yields a list of ``peaks,'' each representing the \textit{m/z} of a specific ion along with an intensity value corresponding to its abundance. Together, this list of peaks is called an ``MS/MS spectrum,'' which serves as a fingerprint of the specific analyte being measured. 
In a typical mass spectrometry run, the instrument will collect on the order of 100,000 such spectra, each corresponding to a distinct peptide (or contaminant).
Canonically, these spectra are then processed by a database search algorithm, with the goal of assigning to each spectrum its generating peptide. 
However, in many settings, there are a variety of other important downstream tasks involving tandem mass spectra, beyond simply assigning the peptide sequence. 
 
As in many other fields, deep learning methods have taken mass spectrometry proteomics by storm. The tasks addressed thus far can be divided into two groups: those that take as input a peptide sequence and those that take as input a spectrum.
In the first category, reviewed by Angelis \textit{et al.}\ \cite{angelis2025peptide}, models predict properties of a given peptide, such as expected fragmentation patterns and retention time, primarily with the goal of improving the sensitivity of database search.   

The second category---tasks that take a spectrum as input---are more relevant to our foundation model.
The most fundamental task in this category is \textit{de novo} peptide sequencing, in which the input is a spectrum and the output is a peptide sequence.
\textit{De novo} sequencing offers an alternative method to database search for solving the spectrum annotation problem without relying on prior knowledge, making it a valuable tool for identifying peptides not present in a pre-defined protein database. 
Algorithms for solving this problem were introduced in the late 1990's \cite{taylor1997sequence} and it was first solved using machine learning in 2015 \cite{ma2015novor}.
Subsequently, DeepNovo \cite{tran2017denovo} combined a convolutional neural network and a recurrent neural network to autoregressively predict the subsequent amino acid when provided an MS/MS spectrum and a peptide prefix.
More recently, Casanovo \cite{yilmaz2022denovo} employed a transformer architecture to frame \textit{de novo} sequencing as a sequence-to-sequence translation task. 
Many methods have since successfully extended Casanovo's transformer architecture to include various ideas such as bidirectional decoding, a contrastive loss, improved positional embeddings, and alternative decoding strategies \cite{bittremieux2024deep}. 

\medskip{\bf Downstream tasks.} We study four downstream tasks that take tandem mass spectra as input: predicting spectrum quality, chimericity, phosphorylation, and glycosylation status.

In spectrum quality prediction, the model is asked to predict whether an observed spectrum is identifiable, meaning that it shows strong and clear signal for a peptide.
This problem has been addressed with a variety of classical machine learning techniques \cite{purvine2004spectral, salmi2006quality, wu2008quality, ma2011scanranker} and more recently using a convolutional neural network \cite{gholamizoj2022speq}.

In mass spectrometry experiments, acquired spectra often inadvertently contain signal from multiple peptides. 
Such spectra are called chimeras, and they can be hard to analyze due to the mixture of signals from each peptide. 
To our knowledge, prediction of chimeric spectra has not previously been solved using machine learning methods.
However, many existing methods generalize the database search procedure to allow for chimeric matches \cite{yu2025msfragger, frejno2024unifying}.

Predicting whether a post-translational modification is present in a given spectrum is another key task of interest. PhoStar uses a random forest to predict whether a given spectrum was generated by a phosphorylated peptide based on a set of hand-designed features \cite{dorl2018phostar}.
AHLF improves on this using a convolutional model which takes as input the full spectrum \cite{altenburg2022ad}. For predicting whether a spectrum contains a peptide which is N- or O-glycosylated, current methods rely on hand-designed rules based on specific fragment ions \cite{sutherland2024autonomous}.

\medskip{\bf Learning representations of spectra.} 
Prior work has investigated learning spectrum representations, but these representations have primarily been used as a dimensionality reduction technique focused specifically on clustering spectra and improving peptide identification. GLEAMS learns low-dimensional spectrum representations optimized such that spectra from the same peptide cluster together \cite{bittremieux2022learned}. Similarly, yHydra co-embeds peptides and spectra such that spectra are close to their generating peptides in embedding space \cite{altenburg2021yhydra}. 

Finally, prior work has investigated foundation models for tandem mass spectra in the metabolomics space. Small molecules typically result in lower complexity mass spectra than peptides, and do not exhibit the same consistent fragmentation patterns along a linear molecular backbone. The methods LSM1-MS2 \cite{asher2024lsm1}, PRISM \cite{healey2024prism}, and DreaMS \cite{bushuiev2024emergence} use unsupervised masked-peak modeling to learn representations of metabolomics mass spectra, demonstrating that these representations improve performance on downstream chemical property prediction tasks. 

\section{\textit{De novo} peptide sequencing as a pre-training task}

To accurately perform \textit{de novo} sequencing, a model needs to capture the fundamental relationships between the analyte present in the instrument (i.e., the peptide) and the observed signal measured by the mass spectrometer. This in turn requires a rich understanding of the physics and chemistry governing peptide chromatography, ionization, and fragmentation. We hypothesize that this fundamental understanding of mass spectra, which is acquired through pre-training on the \textit{de novo} sequencing task, will generalize to other tasks involving mass spectra for which less training data is available. 

Typically, foundation models are trained in an unsupervised manner, so as to benefit from massive datasets of unlabeled training examples. 
However, unlike the settings of natural language processing and computer vision, where there are orders of magnitude more unlabeled training examples than labeled samples, typically 40--60\% of the acquired spectra can be annotated in a given mass spectrometry run and hence can be labeled with their generating peptide. 
Additionally, this labeling is fully automated and high-throughput, with no need for costly human annotations. 
Thus, here we consider making use of these labels to explore the supervised task of \textit{de novo} peptide sequencing as pre-training for a foundation model. 

In this work we perform experiments with a state-of-the art, transformer-based \textit{de novo} sequencing model, Casanovo \cite{yilmaz2022denovo, yilmaz2024sequence}. Casanovo is trained on a dataset of  30 million high-quality labeled tandem mass spectra from the MassIVE-KB spectral library \cite{wang2018assembling}. We use Casanovo's pre-trained spectrum encoder off the shelf as a foundation model for mass spectrometry proteomics. Casanovo uses a standard transformer encoder architecture, where spectra are treated as a sequence of peaks, and each peak is embedded with a positional \textit{m/z} embedding and a learned intensity embedding. This setup allows the model to easily attend to pairs of peaks with specified mass shifts, which is key to interpreting mass spectra. 
To obtain an overall spectrum representation we take the mean of the individual peak embeddings from the spectrum encoder, yielding a single embedding for the spectrum as a whole. To apply the encoder to each downstream task, we train a small task-specific dense predictor head that takes these frozen spectrum embeddings from Casanovo as input.

\section{Downstream tasks}
\begin{figure*}
  \centering
  \begin{tabular}{ll}
    \raisebox{-1\height}{\makebox[0pt][l]{\textbf{(A)}}}%
    \includegraphics[height=2.2in]{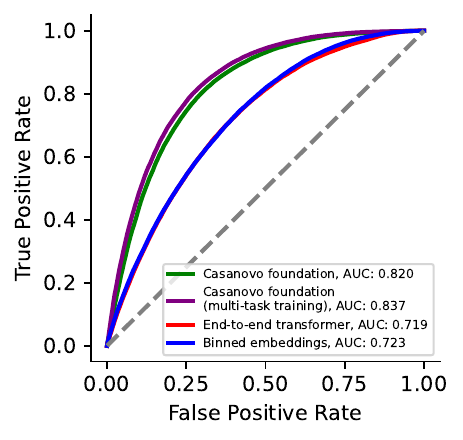} &
    \raisebox{-1\height}{\makebox[0pt][l]{\textbf{(B)}}}%
    \includegraphics[height=2.2in]{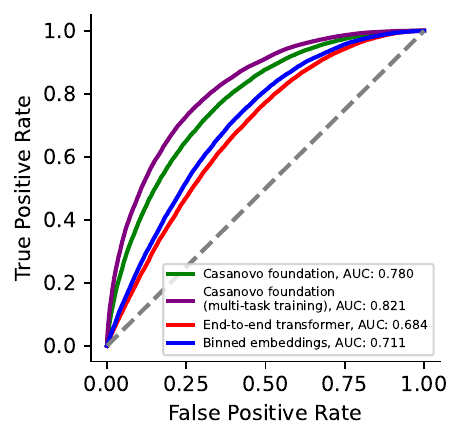} \\
    \raisebox{-1\height}{\makebox[0pt][l]{\textbf{(C)}}}%
    \includegraphics[height=2.2in]{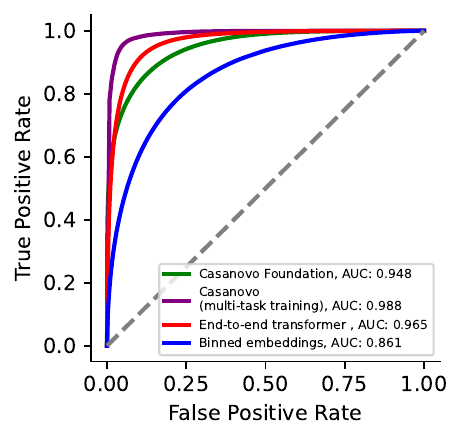} &
    \raisebox{-1\height}{\makebox[0pt][l]{\textbf{(D)}}}%
    \includegraphics[height=2.2in]{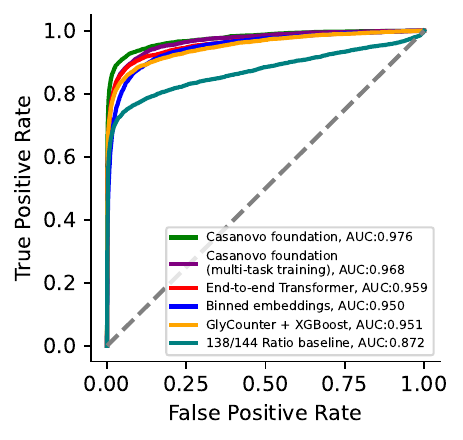} 
  \end{tabular}
  \caption{{\bf Benchmarking Casanovo Foundation on downstream tasks.}
    ROC curves and the area under the curve (AUC) reported for: (A) Spectrum quality prediction, (B) Chimericity prediction, (C) phosphorylation detection, and (D) glycosylation status prediction.
  }
  \label{fig:downstream}
\end{figure*}

We use our foundation model as a starting point to address three downstream tasks.
In each task, we compare the frozen Casanovo encoder coupled with a small task-specific dense predictor head (``Casanovo Foundation'') against at least two baselines.
First, we bin spectrum peaks along the \textit{m/z} axis to obtain spectrum embeddings and then train a gradient boosted decision tree classifier directly on those embeddings (``binned embedding'').
Second, we train a transformer spectrum encoder, which has the same architecture as Casanovo, along with a multilayer preceptron (MLP) classifier head from scratch to learn the downstream tasks end-to-end (``end-to-end transformer''). 
For one of the tasks (phosphorylation detection), we also benchmark against a task-specific state-of-the-art classifier \cite{altenburg2022ad}. For the spectrum quality task, model weights for SPEQ \cite{gholamizoj2022speq}, the current state-of-the-art deep learning method, are unfortunately not publicly available. However, our end-to-end transformer method serves as a conceptually similar baseline representing deep learning models trained directly on the task which take full spectra as input. 

\subsection{Spectrum quality prediction}
The first downstream task we consider is spectrum quality prediction. 
For this task, the goal is to predict whether a given observed MS/MS spectrum will be successfully annotated by database search.
The motivation for this task is three-fold.
First, if we can quickly identify low-quality spectra, then we can save time and potentially boost our statistical power by eliminating these spectra prior to the database search procedure.
Second, spectra that are deemed to be high-quality by the trained model but nonetheless fail to be identified during the database search procedure are good candidates for more expensive computational analyses to find identifications outside of the database.
Finally, by predicting in real time which spectra can be annotated, it is possible to better allocate instrument time towards these analytes. 

To create a labeled dataset for this task, we first randomly sample from the MassIVE repository 20 human mass spectrometry runs that use high-resolution instruments, and we select 10/5/5 files to create training/validation/test splits, where each split contains approximately 450k, 245k and 295k spectra, respectively. 
Spectra that are matched to a peptide under 1\% false discovery rate (FDR) by database search are labeled as high quality, whereas spectra that failed to be matched are annotated as low quality for the binary classification task.
In our dataset, we observe a 40\%/60\% distribution of high- and low-quality spectra. 

Because spectrum quality prediction is a binary classification task and the task is roughly balanced, we use the area under the ROC curve (AUROC) as the primary performance measure.
The presence of foreign spectra (i.e., spectra generated by peptides that are not in the given database, due to contamination or unexpected genetic variation) make this task particularly challenging, because these may be high-quality spectra that will never be confidently assigned a peptide by the database search procedure.
Additionally, because identifications were determined at a 1\% FDR threshold, a small proportion of spectra in the positive class may be incorrect identifications of low-quality spectra. 
As a result, we expect the training and test labels to be fairly noisy for this task and we do not expect \textit{a priori} to be able to achieve AUROC values close to 1.

Applying Casanovo Foundation to this task, we achieve an AUROC of 0.820, outperforming our task-specific end-to-end transformer and the binned embedding baselines (AUROC of 0.719 and 0.723, respectively) (Figure~\ref{fig:downstream}A).
This result suggests that the pre-trained spectrum representations from Casanovo capture properties that are hard to learn from the  quality prediction task alone.
This is not too surprising, given that the \textit{de novo} sequencing pre-training task is both an inherently richer task and took advantage of more data. 

\subsection{Spectrum chimericity prediction}
Tandem mass spectrometry experiments are designed to attempt to isolate individual peptide species, by first separating them by hydrophobicity in the liquid chromatography step and then separating peptides by \textit{m/z} in the first round of mass spectrometry analysis.
Nonetheless, in many cases, two peptides with similar hydrophobicities and \textit{m/z} values end up being fragmented simultaneously.
The result is an MS/MS spectrum that contains peaks corresponding to both peptides.
Such chimeric spectra are difficult to analyze.
Most database search algorithms assign at most one peptide to each spectrum, and even assigning a single peptide to a chimeric spectrum is challenging due to the presence of unexplained peaks from the undetected peptide.

Accordingly, our second downstream task involves detecting when more than one peptide species is responsible for generating a given MS/MS spectrum, i.e., predicting whether it is chimeric or not. Many existing methods generalize the database search procedure to allow chimeric matches \cite{yu2025msfragger, frejno2024unifying}; however, prediction of chimeric spectra has not previously been solved using machine learning methods.
Such a predictor would be useful, for example, in deciding which spectra to provide as input to one of the tools above or in adjusting the settings of an instrument to avoid unwanted chimeras.

To train a chimericity predictor, we use spectra from human, mouse, and yeast samples for training, validation, and test, respectively.
Database search is performed using the wide-window setting in FragPipe \cite{polasky2023msfragger}, which allows spectra to be assigned multiple peptides.
For the binary classification task, spectra assigned more than one peptide are labeled chimeric and spectra annotated with a single peptide are labeled non-chimeric.
Unannotated spectra are discarded. 
For each of the splits, we have roughly 60 thousand spectra annotated with at least one peptide, and approximately 45\% of these spectra are chimeric. Similar to the quality prediction task, we expect that there is noise in these labels due to both false positive and false negative annotations from database search, so achieving close to perfect performance is unlikely. 

Like the quality prediction task, chimericity prediction is a binary classification task without a pronounced class imbalance, so we use AUROC as our primary performance measure.
Comparing Casanovo Foundation to the baseline methods, we again see that it achieves improved performance (AUROC 0.780) compared to the two baselines (AUROC of 0.684 and 0.711) (Figure~\ref{fig:downstream}B). 

\subsection{Post-translational modification detection}
The final type of downstream task we consider is the detection of spectra generated by peptides containing PTMs.
A PTM is a molecular group that attaches to the side-chain of one of the amino acids in a peptide.
The most commonly studied PTMs include phosphorylation, glycosylation, and methylation, but many more potential types of PTMs exist in nature, and some are quite rare.
The peptide database used during database search of MS/MS data can be augmented to include PTMs, but because of the many potential types of PTMs and the fact that a single peptide can harbor multiple PTMs, accounting for all possible modifications is not computationally or statistically feasible.
Additionally, identifying peptides containing PTMs and localizing the modification, i.e., determining which residue the modification is attached to, often requires adjusting settings in the mass spectrometer, such as the fragmentation type or collision energy, to be optimized specifically for that PTM.
Thus, a model capable of identifying which PTMs are associated with a given MS/MS spectrum would be valuable in guiding how data is both collected and subsequently analyzed. 
In fact, simple methods for solving this task are regularly employed in practice to improve the sensitivity and quantitative accuracy of experiments targeting peptides carrying a specific PTM \cite{sutherland2024autonomous, liu2021strategies}. 
Here, we train classifiers to recognize two common types of PTMs.

\begin{table}
    \centering
    \caption{\textbf{Phosphorylation task performance.} Comparison of AHLF, the end-to-end transformer baseline, Casanovo Foundation, and multi-task trained Casanovo Foundation across phosphorylation detection datasets. The 25 datasets listed correspond to the holdout split \emph{a} described in \cite{altenburg2022ad}.  The first two columns indicate the number of non-phosphorylated versus phosphorylated spectra in each dataset. The reported performance metrics are F$_1$ score and AUROC. The best performance on each metric in each row is indicated in bold. AHLF results are directly taken from the paper and thus are only available to two significant figures as originally reported.}
    \label{tab:ahlf}
    \smallskip
    \def\arraystretch{1}
    \resizebox{\textwidth}{!}{ 
\begin{tabular}{ccccccccccc}
\toprule
\multirow{2}{*}{\textbf{Dataset}} & \multirow{2}{*}{\textbf{\begin{tabular}[c]{@{}c@{}}Number \\ non-phospho\end{tabular}}} & \multirow{2}{*}{\textbf{\begin{tabular}[c]{@{}c@{}}Number \\ phospho\end{tabular}}} & \multicolumn{2}{c}{\textbf{AHLF}} & \multicolumn{2}{c}{\textbf{\begin{tabular}[c]{@{}c@{}}End-to-end \\ Transformer\end{tabular}}} & \multicolumn{2}{c}{\textbf{\begin{tabular}[c]{@{}c@{}}Casanovo \\ Foundation\end{tabular}}} & \multicolumn{2}{c}{\textbf{\begin{tabular}[c]{@{}c@{}}Casanovo Foundation \\ (multimodal training)\end{tabular}}} \\ \cmidrule(lr){4-5}\cmidrule(lr){6-7}\cmidrule(lr){8-9}\cmidrule(lr){10-11}
                                  &                                                                                         &                                                                                     & F1          & AUROC               & F1                                             & AUROC                                         & F1                                       & AUROC                                            & F1                                                      & AUROC                                                   \\ \midrule
OVAS                              & 90936                                                                                   & 37720                                                                               & 0.92        & \textbf{0.99}       & 0.878                                          & 0.983                                         & 0.887                                    & 0.982                                            & \textbf{0.967}                                          & \textbf{0.997}                                          \\
TOV-21-Primary                    & 62350                                                                                   & 26978                                                                               & 0.92        & \textbf{0.99}       & 0.872                                          & 0.981                                         & 0.884                                    & 0.982                                            & \textbf{0.968}                                          & \textbf{0.997}                                          \\
ES2-Primary                       & 16297                                                                                   & 6667                                                                                & 0.91        & \textbf{0.99}       & 0.818                                          & 0.981                                         & 0.834                                    & 0.979                                            & \textbf{0.947}                                          & \textbf{0.996}                                          \\
Daudi                             & 150915                                                                                  & 210916                                                                              & 0.90        & 0.96                & 0.953                                          & 0.986                                         & 0.924                                    & 0.980                                            & \textbf{0.975}                                          & \textbf{0.994}                                          \\
U2OS                              & 92329                                                                                   & 205353                                                                              & 0.90        & 0.95                & 0.913                                          & 0.938                                         & 0.909                                    & 0.964                                            & \textbf{0.978}                                          & \textbf{0.992}                                          \\
HaCaT                             & 19216                                                                                   & 113775                                                                              & 0.95        & 0.93                & 0.939                                          & 0.950                                         & 0.964                                    & 0.978                                            & \textbf{0.986}                                          & \textbf{0.993}                                          \\
HT-29                             & 1625                                                                                    & 27531                                                                               & 0.97        & 0.92                & 0.990                                          & \textbf{0.998}                                & 0.973                                    & 0.988                                            & \textbf{0.993}                                          & \textbf{0.998}                                          \\
HeLa                              & 1469194                                                                                 & 2949614                                                                             & 0.89        & 0.92                & 0.923                                          & 0.953                                         & 0.917                                    & 0.955                                            & \textbf{0.973}                                          & \textbf{0.990}                                          \\
HEPG2                             & 426                                                                                     & 45416                                                                               & 0.98        & 0.92                & 0.989                                          & 0.998                                         & 0.971                                    & 0.983                                            & \textbf{0.995}                                          & \textbf{0.999}                                          \\
A549                              & 4068                                                                                    & 172792                                                                              & 0.92        & 0.91                & 0.977                                          & 0.988                                         & 0.950                                    & 0.960                                            & \textbf{0.981}                                          & \textbf{0.991}                                          \\
Colon                             & 8359                                                                                    & 28798                                                                               & 0.90        & 0.90                & \textbf{0.938}                                 & 0.965                                         & 0.904                                    & 0.925                                            & 0.937                                                   & \textbf{0.971}                                          \\
Primary-Gastro                    & 22026                                                                                   & 219767                                                                              & 0.94        & 0.88                & 0.931                                          & 0.929                                         & 0.910                                    & 0.929                                            & \textbf{0.971}                                          & \textbf{0.981}                                          \\
LNCaP                             & 53851                                                                                   & 5200                                                                                & 0.45        & 0.87                & 0.692                                          & \textbf{0.987}                                & 0.346                                    & 0.877                                            & \textbf{0.716}                                          & 0.980                                                   \\
RPMI-8226                         & 1184                                                                                    & 413                                                                                 & 0.65        & 0.87                & \textbf{0.992}                                 & \textbf{0.999}                                & 0.727                                    & 0.938                                            & 0.907                                                   & 0.985                                                   \\
HEK293                            & 322811                                                                                  & 332690                                                                              & 0.73        & 0.86                & \textbf{0.910}                                 & \textbf{0.966}                                & 0.760                                    & 0.837                                            & 0.894                                                   & 0.952                                                   \\
Primary-Prostate                  & 9223                                                                                    & 100617                                                                              & 0.89        & 0.86                & 0.954                                          & 0.982                                         & 0.894                                    & 0.919                                            & \textbf{0.979}                                          & \textbf{0.990}                                          \\
Primary-AML                       & 494184                                                                                  & 5893                                                                                & 0.29        & 0.85                & 0.755                                          & \textbf{0.988}                                & 0.583                                    & 0.945                                            & \textbf{0.785}                                          & 0.984                                                   \\
Kasumi-1                          & 2294                                                                                    & 29470                                                                               & 0.88        & 0.85                & \textbf{0.995}                                 & \textbf{0.998}                                & 0.902                                    & 0.844                                            & 0.969                                                   & 0.985                                                   \\
HPAC                              & 594                                                                                     & 807                                                                                 & 0.56        & 0.78                & 0.962                                          & 0.985                                         & 0.733                                    & 0.820                                            & \textbf{0.991}                                          & \textbf{0.994}                                          \\
SU.86.86                          & 909                                                                                     & 984                                                                                 & 0.53        & 0.77                & \textbf{0.967}                                 & \textbf{0.996}                                & 0.664                                    & 0.782                                            & 0.769                                                   & 0.943                                                   \\
CFPAC-1                           & 999                                                                                     & 780                                                                                 & 0.52        & 0.76                & 0.714                                          & 0.897                                         & 0.600                                    & 0.803                                            & \textbf{0.755}                                          & \textbf{0.933}                                          \\
PANC-05-04                        & 1079                                                                                    & 1426                                                                                & 0.55        & 0.74                & \textbf{0.959}                                 & \textbf{0.993}                                & 0.685                                    & 0.78                                             & 0.789                                                   & 0.925                                                   \\
PANC-02-03                        & 273                                                                                     & 815                                                                                 & 0.56        & 0.72                & 0.671                                          & 0.896                                         & 0.731                                    & 0.819                                            & \textbf{0.791}                                          & \textbf{0.927}                                          \\
OVSAYO                            & 11515                                                                                   & 28                                                                                  & 0.02        & 0.69                & 0.021                                          & 0.825                                         & 0.026                                    & \textbf{0.868}                                   & \textbf{0.052}                                          & 0.821                                                   \\
HDMVEC                            & 4320                                                                                    & 2961                                                                                & 0.40        & 0.60                & \textbf{0.873}                                 & \textbf{0.970}                                & 0.618                                    & 0.783                                            & 0.844                                                   & 0.937                                                  
\\ \bottomrule \end{tabular}
}
\end{table}

\subsubsection{Phosphorylation detection}
We first consider the detection of spectra from phosphorylated peptides. Protein phosphorylation is arguably one of the most important and well studied PTMs, driving key physiological activities such as energy metabolism, cell proliferation and growth, apoptosis, and signal transduction \cite{ardito2017crucial}. We frame phosphorylation prediction as a binary classification task, predicting whether or not a given spectrum derives from a phosphorylated peptide.

To train a classifier, we use 19.2 million labeled spectra from the human phosphoproteome dataset \cite{ochoa2020functional} which were used to train AHLF \cite{altenburg2022ad}, a state-of-the-art phosphorylation predictor. 
The human phosphoproteome consists of 112 individual PRIDE datasets, containing 101 human cell or tissue types, where each dataset was collected with phospho-enrichment assays.
To create labeled data for training AHLF, the human phosphoproteome was subjected to database search, and a binary label was assigned to spectra indicating phosphorylated or unphosphorylated peptides (see \cite{altenburg2022ad} for details).
Of the resulting 19.2 million labeled spectra, 54\% are phosphorylated.
Following the cross-validation setup described in \cite{altenburg2022ad}, we use the same train, validation, and test splits as the AHLF-$\alpha$ model.
For phosphorylation prediction, we use the F$_1$ score in addition to the AUROC metric to account for class imbalances at the level of individual datasets within the test set.

Comparing the ROC curves for Casanovo Foundation to our two baselines for this task, we observe that, unlike the previous two tasks, our foundation model (AUROC 0.948) performs worse than the end-to-end transformer model (AUROC 0.965) but better than the binned spectrum baseline (AUROC 0.861) (Figure~\ref{fig:downstream}C). Breaking performance down across each of the 25 test datasets and comparing to AHLF results, we observe that Casanovo Foundation performs somewhat better than AHLF, yielding higher performance on 19/25 datasets for each metric. However, the end-to-end transformer baseline outperforms both, with a higher F1 score on 19/25 and a higher AUROC on 17/25 datasets (Table~\ref{tab:ahlf}). This result is not too surprising, because foundation modeling is not expected to provide a major advantage on tasks with very large amounts of high-quality labeled data available for training. Furthermore, the dataset used for pre-training Casanovo does not contain phosphopeptides, meaning that the model may not have learned to fully recognize the importance of specific peaks and mass shifts which are indicative of phosphorylation. 

\begin{figure}
\centering
  \begin{subfigure}[c]{.61\linewidth}
  \subfigimg[width=1.08\linewidth]{\textbf{(A)}}{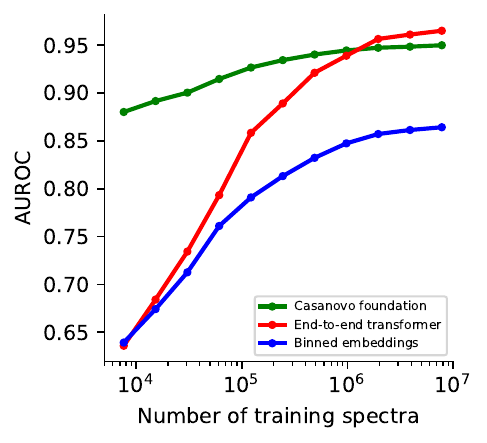}
  \end{subfigure}\hfill
  \begin{tabular}[c]{@{}c@{}}
    \begin{subfigure}[c]{.3\linewidth}
      \subfigimg[width=\linewidth]{\textbf{(B)}}{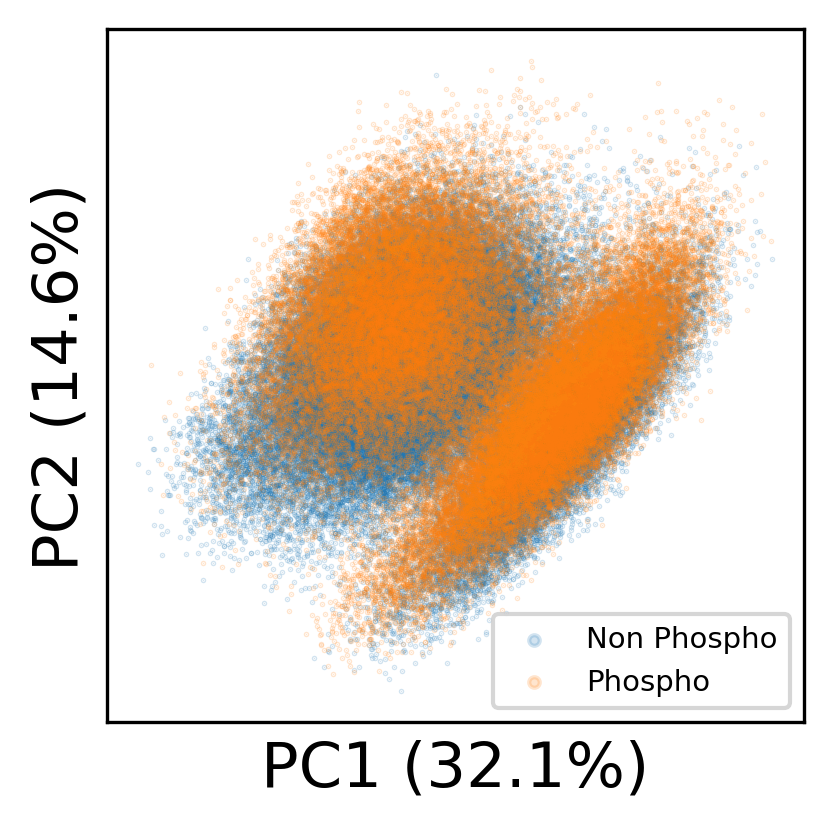}
    \end{subfigure}\\
    \begin{subfigure}[c]{.3\linewidth}
       \subfigimg[width=\linewidth]{\textbf{(C)}}{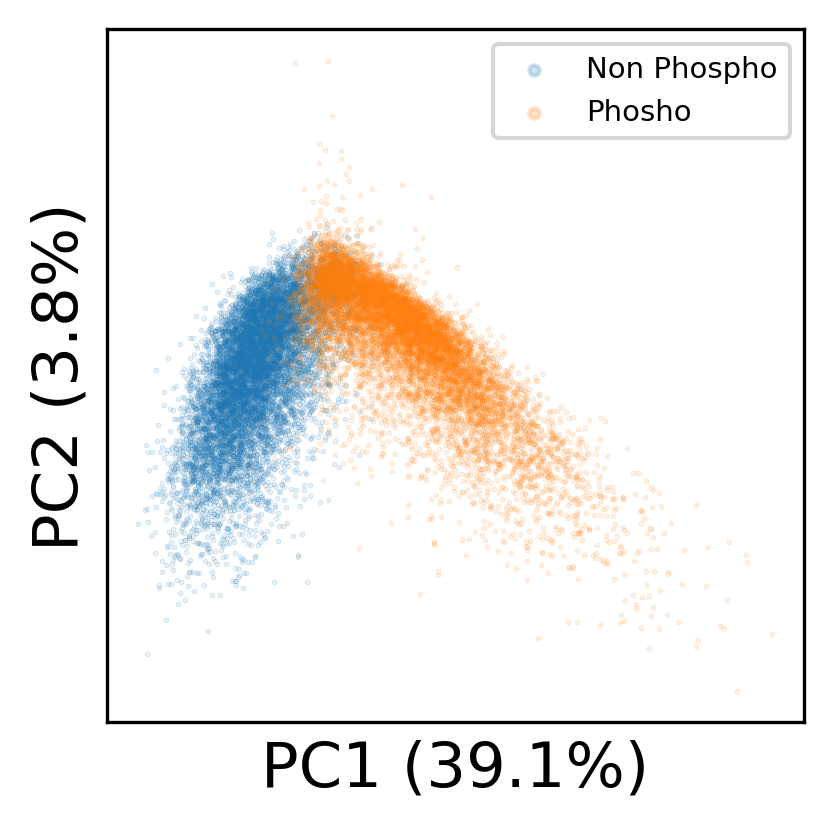}
    \end{subfigure}
  \end{tabular}
\caption{{\bf Phosphorylation task experiments}. (A) Learning curve showing the performance of Casanovo Foundation and baselines on the phosphorylation prediction task when trained on datasets of varying size. 
(B) PCA plot visualizing the spectrum embeddings from  the pre-trained encoder for spectra in the phosphorylation task test set. (C) Embeddings for the same spectra after multi-task fine-tuning. The percentage of variance explained by each component is indicated in parentheses.}
\label{fig:learning_curve}
\end{figure}

For PTMs other than phosphorylation, which may be both rarer biologically and lack well-established enrichment protocols, such a large training set is unavailable. For these modifications, we reasoned that the foundation modeling approach may prove more valuable. Accordingly, to investigate the relationship between the number of available training samples and the performance of each model, we create a series of 10 nested subsets of the phosphoproteomics training data, which range in size from roughly 7,700 to 7.7 million training spectra. The relative performance of foundation modeling to our supervised baselines on each subset informs in what settings foundation modeling may provide an advantage. We find that for datasets with fewer than $\sim$1 million spectra, the performance of our Casanovo Foundation model and the end-to-end transformer baseline cross over, with the foundation model showing the best performance (Figure~\ref{fig:learning_curve}A). The difference in performance grows as the size of the training set decreases.  Strikingly, Casanovo Foundation achieves an AUROC of 0.881 when trained on a dataset of just 7,615 spectra, compared to 0.639 for the binned embedding baseline and 0.635 for the end-to-end transformer baseline. Given that for many rarer or less-well-studied PTMs, assembling a dataset of even $\sim$100,000 spectra may prove a significant challenge, this result suggests the potential utility of Casanovo Foundation for other PTM prediction tasks. 

\subsubsection{Glycosylation determination.}
To explore the task of PTM prediction in a setting where foundation modeling may be more necessary, we turn to another important modification for which training data is less readily available. 
Protein glycosylation is a complex PTM where various combinations of mono- and oligosaccharides are attached to specific residues. 
Here, we consider the task of predicting the glycosylation class of a peptide from its spectrum. 
The two most common classes of glycosylation are N-glycosylation, where glycans are attached to the nitrogen side chain of asparagine residues within a specific motif, and O-glycosylation, where glycans are attached to hydroxyl groups of serine and threonine residues \cite{he2024glycosylation, schwarz2011mechanisms, li2022recent, wandall2021global}.
Recognizing whether a given spectrum represents a glycosylated peptide is straightforward due to the presence of characteristic oxonium ions that are generated upon collisional activation \cite{wu2014novel, saba2012increasing, singh2012higher, zhao2011combining}.
However, distinguishing N-glycosylation from O-glycosylation is more difficult (Supplementary note \ref{sec:glyco_note}). 

Effectively classifying glycopeptides, especially N- vs O-glycopeptides, is critical because the optimal dissociation method differs for N- and O-glycopeptides. Tryptic N-glycopeptides typically only have a single potential glycosite, meaning that higher-energy collisional dissociation (HCD) is sufficient for both identifying and localizing N-glycosylation \cite{riley2020optimal}. On the other hand, O-glycopeptide sequences can often contain multiple potential O-glycosites per peptide, and thus require the collection of alternative dissociation methods, e.g., electron-transfer dissociation (ETD), to generate peptide fragment ions that retain glycan modifications that facilitate localization. Acquiring ETD spectra incurs a significant overhead in instrument time. Thus, by predicting whether a given HCD spectrum contains an N- versus an O-glycopeptide, we can intelligently guide the data acquisition to spend instrument time acquiring ETD spectra only for the precursor ions for which it is necessary \cite{sutherland2024autonomous}. 

To train a model to distinguish N- versus O-glycsolyation, we use a publicly available dataset of the mouse brain glycoproteome produced by DQGlyco.
This dataset contains 252,970 total glycopeptide identifications, of which 25,757 (10.2\%) are O-glycosylated \cite{potel2025uncovering}
In addition to our two standard baselines, for this task we also consider two domain-specific baselines. The first looks at the ratio in intensity between the oxonium ion at 138 \textit{m/z} to that at 144 \textit{m/z}. This ratio between the abundances of expected product ions from N- and O-glycans is currently used in practice for real-time prediction in glycoproteomics experiments \cite{sutherland2024autonomous}. The second baseline is a slightly more sophisticated version of the prior approach, which trains an XGBoost classifier on the abundance of these two oxonium ions, along with the abundances of 52 other oxonium ions, extracted by GlyCounter \cite{kothlow2025extracting}, that are known to be characteristic of glycosylation.

Evaluating the performance of each method, we find that the domain-specific baselines are already reasonably good, with an AUROC of 0.872 for the 138/144 ratio and 0.951 for GlyCounter+XGBoost. The binned embedding baseline and end-to-end transformer baselines perform similarly, achieving AUROCs of 0.950 and 0.959, respectively. However, we again find that Casanovo Foundation offers the best results, achieving an AUROC of 0.976 (Figure~\ref{fig:downstream}D). Given the significant class imbalance in the data, with only $\sim$10\% of the data coming from the positive class, we also plot precision-recall curves for this task (Supplementary Figure~\ref{fig:glyco_pr}). This accentuates the difference in performance between methods, with Casanovo Foundation achieving a AUPR of 0.914, compared to 0.753, 0.860, 0.811, and 0.867 for the ratio, GlyCounter, binned, and transformer baselines, respectively. 

\section{Multi-task training}
Having demonstrated the utility of pre-trained spectrum representations from Casanovo Foundation on various downstream tasks, we next turn to strategies for improving the representations further. 
To this end, we fine-tune our pre-trained Casanovo encoder using a multi-task learning strategy. In this setup, we add three task-specific prediction heads on top of the shared pre-trained spectrum encoder. 
The model is then jointly optimized on the spectrum quality, chimericity, and phosphorylation prediction downstream tasks, in addition to the main \textit{de novo} sequencing task used during pre-training. 
During each training step, the multi-task model receives one batch of spectra from each task and minimizes their summed loss.
To approximately balance the amount of training data from each of the three downstream tasks, we use the 1/32 downsampled phosphorylation dataset (243,710 spectra) from the above learning curve experiment for training.
We hypothesize that the diversity in training data and tasks during joint training will introduce the model to a broader distribution of spectra than was seen during pre-training, thereby helping the encoder to recognize and extract a wider range of important spectrum features. 
Specifically, during \textit{de novo} sequencing pre-training, the model saw no low-quality spectra nor spectra with phosphorylation as a PTM. 

Having trained our multi-task encoder, we evaluate it using the same procedure as described in Section 4. For each downstream task, we obtain spectral embeddings from the pre-trained encoder and train a task-specific classifier directly on these representations. We find that our multi-task training improves performance on all three tasks, although to varying degrees: spectrum quality prediction improves from an AUROC of 0.820 to 0.837, chimericity prediction improves from 0.780 to 0.821, and phosphorylation prediction improves from 0.948 to 0.988 (Figures~\ref{fig:downstream}A-C). This performance on the phosphorylation task now surpasses the task-specific transformer model trained on the full phosphorylation dataset, achieving the best performance out of all methods for 17 of the 25 individual datasets in the test set (Table~\ref{tab:ahlf}). Notably, we observe that our phosphorylation classifier trained on multi-task encoder embeddings converges very quickly, unlike the end-to-end transformer baseline, which continues to improve when trained on larger datasets all the way up to the full 7.7 million spectra. Thus, we opted to use the same downsampled dataset for training the phosphorylation predictor head that was used for pre-training the multi-task encoder. This means that not only does Casanovo Foundation with multi-task training yield state-of-the-art performance on the phosphorylation prediction task, but it does so while using, in total, only 1/32 times as much training data as competing methods. 

To visualize the structure of the latent space learned by our spectrum encoders and to better understand how multi-task training improves performance, we performed principal component analysis (PCA) of the spectrum embeddings of spectra from each of the three tasks. For the pretrained encoder, we observed a heavy overlap between spectra from each class for all three tasks (Figure~\ref{fig:learning_curve}A, Supplementary Figures~\ref{fig:embeddings}A--B). This overlap indicates that the features relevant to each task are not assigned high weight by the encoder. After joint training, however, the spectrum representations from the phosphorylation dataset show clear separation in the first principal component (Figure~\ref{fig:learning_curve}D), which explains why the task-specific prediction head trained on these representations requires so little training data to achieve good performance. For the other two datasets there is still considerable overlap between the embeddings of each class, as may be expected given the comparatively lower AUROCs achieved on these tasks. However, we still observe much clearer separation than is seen for the embeddings without multi-task fine-tuning (Supplementary Figures~\ref{fig:embeddings}C--D). 

Having shown that multi-task finetuning of the spectrum encoder improves spectrum representations for the tasks included in training, we next sought to evaluate whether this approach also improves performance for other tasks \textit{not} included in training. Thus, we evaluated the representations from the multi-task encoder on the unseen glycosylation status prediction task above. We find that the representations learned by the multi-task model are less useful for this task, achieving an AUROC of 0.968, compared to the 0.976 achieved by the non-finetuned encoder. This suggests that multi-task training does not improve performance on other downstream tasks which are not included in the training, at least for the single downstream task which we tested here. 

Overall, these results indicate that the greatest performance on a given task is obtained when the pre-trained Casanovo spectrum encoder is fine-tuned on that task in a multi-task training setup. However, the benefits of this fine-tuning are task-specific and do not necessarily generalize to new tasks. Additionally, the benefits of this approach come at the expense of introducing complexity and significantly more computational cost, which may not be worthwhile for many use cases. 

\section{Conclusion and future work}

In this work, we demonstrate that the spectrum encoder learned by a model trained on the \textit{de novo} sequencing task is generally applicable as a foundation model for tandem mass spectrometry data. Small models trained on frozen spectrum embeddings give good performance across a wide range of downstream tasks, and multi-task fine-tuning of the spectrum encoder improves performance further, with Casanovo Foundation ultimately achieving state-of-the-art performance on all downstream tasks it was applied to. These results demonstrate the utility of foundation models for mass spectrometry proteomics as a flexible starting point for solving novel tasks without the need for massive task-specific labeled datasets. 

One promising avenue for future research is to replace or augment the \textit{de novo} pre-training with an unsupervised pre-training task, as has been done in metabolomics \cite{asher2024lsm1, healey2024prism, bushuiev2024emergence}.
Although this will not dramatically increase the training dataset size, it may lead to richer and more generalizable spectrum representations.
Additionally, such an approach would allow the inclusion of more diverse spectra, including those not readily annotatable by database search.



\bibliographystyle{plain}
\bibliography{refs}

\begin{thebibliography}{10}

\bibitem{aebersold:mass-spectrometric}
R.~Aebersold and M.~Mann.
\newblock Mass-spectrometric exploration of proteome structure and function.
\newblock {\em Nature}, 537:347--355, 2016.

\bibitem{altenburg2021yhydra}
T.~Altenburg, T.~Muth, and B.~Y. Renard.
\newblock {{yHydra}}: {{Deep Learning}} enables an {{Ultra Fast Open Search}} by {{Jointly Embedding MS}}/{{MS Spectra}} and {{Peptides}} of {{Mass Spectrometry-based Proteomics}}, December 2021.

\bibitem{altenburg2022ad}
Tom Altenburg, Sven~H. Giese, Shengbo Wang, Thilo Muth, and Bernhard~Y. Renard.
\newblock Ad hoc learning of peptide fragmentation from mass spectra enables an interpretable detection of phosphorylated and cross-linked peptides.
\newblock {\em Nature Machine Intelligence}, 4(4):378--388, April 2022.
\newblock Publisher: Nature Publishing Group.

\bibitem{angelis2025peptide}
J.~Angelis, E.~A. Schr{\"o}der, Z.~Xiao, W.~Gabriel, and M.~Wilhelm.
\newblock Peptide property prediction for mass spectrometry using {AI}: An introduction to state of the art models.
\newblock {\em Proteomics}, page e202400398, 2025.

\bibitem{ardito2017crucial}
Fatima Ardito, Michele Giuliani, Donatella Perrone, Giuseppe Troiano, and Lorenzo~Lo Muzio.
\newblock The crucial role of protein phosphorylation in cell signaling and its use as targeted therapy ({Review}).
\newblock {\em International Journal of Molecular Medicine}, 40(2):271--280, August 2017.

\bibitem{asher2024lsm1}
Gabriel Asher, Jennifer~M. Campbell, Jack Geremia, and Timothy Kassis.
\newblock {LSM1}-{MS2}: {A} {Self}-{Supervised} {Foundation} {Model} for {Tandem} {Mass} {Spectrometry} {Applications}, {Encompassing} {Extensive} {Chemical} {Property} {Predictions} and {Spectral} {Matching}, February 2024.

\bibitem{bittremieux2024deep}
W.~Bittremieux, V.~Ananth, W.~Fondrie, C.~Melendez, M.~Pominova, J.~Sanders, B.~Wen, M.~Yilmaz, and W.~S. Noble.
\newblock Deep learning methods for de novo peptide sequencing.
\newblock {\em Mass Spectrometry Reviews}, 2024.
\newblock In press.

\bibitem{bittremieux2022learned}
W.~Bittremieux, D.~H. May, J.~Bilmes, and W.~S. Noble.
\newblock A learned embedding for efficient joint analysis of millions of mass spectra.
\newblock {\em Nature Methods}, 19(6):675--678, 2022.

\bibitem{bommasani2021opportunities}
R.~Bommasani, D.~A. Hudson, E.~Adeli, R.~Altman, S.~Arora, S.~{von Arx}, M.~S. Bernstein, J.~Bohg, A.~Bosselut, and E.~Brunskill.
\newblock On the opportunities and risks of foundation models.
\newblock {\em arXiv preprint arXiv:2108.07258}, 2021.

\bibitem{bushuiev2024emergence}
Roman Bushuiev, Anton Bushuiev, Raman Samusevich, Corinna Brungs, Josef Sivic, and Tom{\'a}{\v{s}} Pluskal.
\newblock Emergence of molecular structures from repository-scale self-supervised learning on tandem mass spectra.
\newblock {\em chemRxiv}, 2024.
\newblock doi:10.26434/chemrxiv-2023-kss3r-v2.

\bibitem{chambers2012cross-platform}
M.~C. Chambers, B.~Maclean, R.~Burke, D.~Amodei, D.~L. Ruderman, S.~Neumann, L.~Gatto, B.~Fischer, B.~Pratt, J.~Egertson, K.~Hoff, D.~Kessner, N.~Tasman, N.~Shulman, B.~Frewen, T.~A. Baker, M.~Y. Brusniak, C.~Paulse, D.~Creasy, L.~Flashner, K.~Kani, C.~Moulding, S.~L. Seymour, L.~M. Nuwaysir, B.~Lefebvre, F.~Kuhlmann, J.~Roark, P.~Rainer, S.~Detlev, T.~Hemenway, A.~Huhmer, J.~Langridge, B.~Connolly, T.~Chadick, K.~Holly, J.~Eckels, E.~W. Deutsch, R.~L. Moritz, J.~E. Katz, D.~B. Agus, M.~J. MacCoss, D.~L. Tabb, and P.~Mallick.
\newblock A cross-platform toolkit for mass spectrometry and proteomics.
\newblock {\em Nature Biotechnology}, 30(10):918--920, 2012.

\bibitem{chen:xgboost}
T.~Chen and C.~Guestrin.
\newblock {XGBoost}: A scalable tree boosting system.
\newblock In {\em Proceedings of the 22nd ACM SIGKDD International Conference on Knowledge Discovery and Data Mining}, KDD '16, pages 785--794, New York, NY, USA, 2016. ACM.

\bibitem{dorl2018phostar}
Sebastian Dorl, Stephan Winkler, Karl Mechtler, and Viktoria Dorfer.
\newblock {PhoStar}: {Identifying} {Tandem} {Mass} {Spectra} of {Phosphorylated} {Peptides} before {Database} {Search}.
\newblock {\em Journal of Proteome Research}, 17(1):290--295, January 2018.
\newblock Publisher: American Chemical Society.

\bibitem{frejno2024unifying}
Martin Frejno, Michelle~T Berger, Johanna Tueshaus, Alexander Hogrebe, Florian Seefried, Michael Graber, Patroklos Samaras, Samia~Ben Fredj, Vishal Sukumar, Layla Eljagh, Igor Bronshtein, Lizi Mamisashvili, Markus Schneider, Siegfried Gessulat, Tobias Schmidt, Bernhard Kuster, Daniel~P Zolg, and Mathias Wilhelm.
\newblock Unifying the analysis of bottom-up proteomics data with chimerys.
\newblock {\em bioRxiv}, 2024.

\bibitem{gholamizoj2022speq}
S.~Gholamizoj and B.~Ma.
\newblock {SPEQ}: quality assessment of peptide tandem mass spectra with deep learning.
\newblock {\em Bioinformatics}, 38(6):1568--1574, 2022.

\bibitem{he2024glycosylation}
Mengyuan He, Xiangxiang Zhou, and Xin Wang.
\newblock Glycosylation: mechanisms, biological functions and clinical implications.
\newblock {\em Signal Transduction and Targeted Therapy}, 9(1):1--33, August 2024.
\newblock Publisher: Nature Publishing Group.

\bibitem{healey2024prism}
David Healey, Daniel Domingo-Fernández, J~Taylor, Christoph Krettler, Rose Lightheart, Tyson Park, Tobias Kind, August Allen, and Viswa Colluru.
\newblock {PRISM}: {A} foundation model for life's chemistry.

\bibitem{kingma2015adam}
D.~Kingma and J.~Ba.
\newblock Adam: A method for stochastic optimization.
\newblock In {\em Proceedings of the 3rd International Conference on Learning Representations}, 2015.

\bibitem{kothlow2025extracting}
Kathryn Kothlow, Haley~M. Schramm, Kayla~A. Markuson, Jacob~H. Russell, Emmajay Sutherland, Tim~S. Veth, Ruby Zhang, Anna~G. Duboff, Vishnu~R. Tejus, Leah~E. McDermott, Laura~S. Dräger, and Nicholas~M. Riley.
\newblock Extracting informative glycan-specific ions from glycopeptide {MS}/{MS} spectra with {GlyCounter}, March 2025.
\newblock Pages: 2025.03.24.645139 Section: New Results.

\bibitem{li2022recent}
Jiajia Li, Bo~Guo, Wenqi Zhang, Shuang Yue, Shan Huang, Song Gao, Junfeng Ma, John~F. Cipollo, and Shuang Yang.
\newblock Recent advances in demystifying {O}-glycosylation in health and disease.
\newblock {\em PROTEOMICS}, 22(23-24):2200156, 2022.
\newblock \_eprint: https://onlinelibrary.wiley.com/doi/pdf/10.1002/pmic.202200156.

\bibitem{liu2021strategies}
Xinyue Liu, Rose Fields, Devin~K. Schweppe, and Joao~A. Paulo.
\newblock Strategies for {Mass} {Spectrometry}-based {Phosphoproteomics} using {Isobaric} {Tagging}.
\newblock {\em Expert review of proteomics}, 18(9):795--807, September 2021.

\bibitem{ma2015novor}
B.~Ma.
\newblock Novor: Real-time peptide de novo sequencing software.
\newblock {\em Journal of the American Society for Mass Spectrometry}, 26:1885--1894, 2015.

\bibitem{ma2011scanranker}
Ze-Qiang Ma, Matthew~C Chambers, Amy-Joan~L Ham, Kristin~L Cheek, Corbin~W Whitwell, Hans-Rudolf Aerni, Birgit Schilling, Aaron~W Miller, Richard~M Caprioli, and David~L Tabb.
\newblock {ScanRanker}: Quality assessment of tandem mass spectra via sequence tagging.
\newblock {\em Journal of Proteome Research}, 10(7):2896--2904, 2011.

\bibitem{maccoss2023sampling}
Michael~J. MacCoss, Javier~Antonio Alfaro, Danielle~A. Faivre, Christine~C. Wu, Meni Wanunu, and Nikolai Slavov.
\newblock Sampling the proteome by emerging single-molecule and mass spectrometry methods.
\newblock {\em Nature methods}, 20(3):339--346, March 2023.

\bibitem{ochoa2020functional}
David Ochoa, Andrew~F. Jarnuczak, Cristina Viéitez, Maja Gehre, Margaret Soucheray, André Mateus, Askar~A. Kleefeldt, Anthony Hill, Luz Garcia-Alonso, Frank Stein, Nevan~J. Krogan, Mikhail~M. Savitski, Danielle~L. Swaney, Juan~A. Vizcaíno, Kyung-Min Noh, and Pedro Beltrao.
\newblock The functional landscape of the human phosphoproteome.
\newblock {\em Nature Biotechnology}, 38(3):365--373, March 2020.

\bibitem{polasky2023msfragger}
Daniel~A Polasky, Daniel~J Geiszler, Fengchao Yu, Kai Li, Guo~Ci Teo, and Alexey~I Nesvizhskii.
\newblock {MSFragger-Labile}: A flexible method to improve labile {PTM} analysis in proteomics.
\newblock {\em Molecular \& Cellular Proteomics}, 22(5), 2023.

\bibitem{potel2025uncovering}
Clément~M. Potel, Mira~Lea Burtscher, Martin Garrido-Rodriguez, Amber Brauer-Nikonow, Isabelle Becher, Cecile Le~Sueur, Athanasios Typas, Michael Zimmermann, and Mikhail~M. Savitski.
\newblock Uncovering protein glycosylation dynamics and heterogeneity using deep quantitative glycoprofiling ({DQGlyco}).
\newblock {\em Nature Structural \& Molecular Biology}, pages 1--16, February 2025.
\newblock Publisher: Nature Publishing Group.

\bibitem{purvine2004spectral}
Samuel Purvine, Natali Kolker, and Eugene Kolker.
\newblock Spectral quality assessment for high-throughput tandem mass spectrometry proteomics.
\newblock {\em Omics: a journal of integrative biology}, 8(3):255--265, 2004.

\bibitem{radford:learning}
Alec Radford, Jong~Wook Kim, Chris Hallacy, Aditya Ramesh, Gabriel Goh, Sandhini Agarwal, Girish Sastry, Amanda Askell, Pamela Mishkin, Jack Clark, et~al.
\newblock Learning transferable visual models from natural language supervision.
\newblock In {\em International Conference on Machine Learning}, pages 8748--8763. PMLR, 2021.

\bibitem{radford2018improving}
Alec Radford, Karthik Narasimhan, Tim Salimans, Ilya Sutskever, et~al.
\newblock Improving language understanding by generative pre-training.
\newblock 2018.

\bibitem{riley2020optimal}
Nicholas~M. Riley, Stacy~A. Malaker, Marc~D. Driessen, and Carolyn~R. Bertozzi.
\newblock Optimal {Dissociation} {Methods} {Differ} for {N}- and {O}-{Glycopeptides}.
\newblock {\em Journal of Proteome Research}, 19(8):3286--3301, August 2020.
\newblock Publisher: American Chemical Society.

\bibitem{saba2012increasing}
Julian Saba, Sucharita Dutta, Eric Hemenway, and Rosa Viner.
\newblock Increasing the {Productivity} of {Glycopeptides} {Analysis} by {Using} {Higher}-{Energy} {Collision} {Dissociation}-{Accurate} {Mass}-{Product}-{Dependent} {Electron} {Transfer} {Dissociation}.
\newblock {\em International Journal of Proteomics}, 2012(1):560391, 2012.
\newblock \_eprint: https://onlinelibrary.wiley.com/doi/pdf/10.1155/2012/560391.

\bibitem{salmi2006quality}
Jussi Salmi, Robert Moulder, Jan-Jonas Filen, Olli~S Nevalainen, Tuula~A Nyman, Riitta Lahesmaa, and Tero Aittokallio.
\newblock Quality classification of tandem mass spectrometry data.
\newblock {\em Bioinformatics}, 22(4):400--406, 2006.

\bibitem{schwarz2011mechanisms}
Flavio Schwarz and Markus Aebi.
\newblock Mechanisms and principles of {N}-linked protein glycosylation.
\newblock {\em Current Opinion in Structural Biology}, 21(5):576--582, October 2011.

\bibitem{singh2012higher}
Charandeep Singh, Cleidiane~G. Zampronio, Andrew~J. Creese, and Helen~J. Cooper.
\newblock Higher {Energy} {Collision} {Dissociation} ({HCD}) {Product} {Ion}-{Triggered} {Electron} {Transfer} {Dissociation} ({ETD}) {Mass} {Spectrometry} for the {Analysis} of {N}-{Linked} {Glycoproteins}.
\newblock {\em Journal of Proteome Research}, 11(9):4517--4525, September 2012.
\newblock Publisher: American Chemical Society.

\bibitem{sutherland2024autonomous}
Emmajay Sutherland, Tim~S. Veth, William~D. Barshop, Jacob~H. Russell, Kathryn Kothlow, Jesse~D. Canterbury, Christopher Mullen, David Bergen, Jingjing Huang, Vlad Zabrouskov, Romain Huguet, Graeme~C. McAlister, and Nicholas~M. Riley.
\newblock Autonomous {Dissociation}-type {Selection} for {Glycoproteomics} {Using} a {Real}-{Time} {Library} {Search}.
\newblock {\em Journal of Proteome Research}, 23(12):5606--5614, December 2024.
\newblock Publisher: American Chemical Society.

\bibitem{taylor1997sequence}
J.~A. Taylor and R.~S. Johnson.
\newblock Sequence database searches via {\em de novo} peptide sequencing by tandem mass spectrometry.
\newblock {\em Rapid Communications in Mass Spectrometry}, 11:1067--1075, 1997.

\bibitem{tran2017denovo}
N.~H. Tran, X.~Zhang, L.~Xin, B.~Shan, and M.~Li.
\newblock De novo peptide sequencing by deep learning.
\newblock {\em Proceedings of the National Academy of Sciences of the United States of America}, 31:8247--8252, 2017.

\bibitem{wandall2021global}
Hans~H. Wandall, Mathias A.~I. Nielsen, Sarah King-Smith, Noortje de~Haan, and Ieva Bagdonaite.
\newblock Global functions of {O}-glycosylation: promises and challenges in {O}-glycobiology.
\newblock {\em The FEBS Journal}, 288(24):7183--7212, 2021.
\newblock \_eprint: https://onlinelibrary.wiley.com/doi/pdf/10.1111/febs.16148.

\bibitem{wang2018assembling}
M.~Wang, J.~Wang, J.~Carver, B.~S. Pullman, S.~W. Cha, and N.~Bandeira.
\newblock Assembling the community-scale discoverable human proteome.
\newblock {\em Cell Systems}, 7:412--421.e5, 2018.

\bibitem{wen2024carafe}
Bo~Wen, Chris Hsu, Wen-Feng Zeng, Michael Riffle, Alexis Chang, Miranda Mudge, Brook~L Nunn, Matthew~D Berg, Judit Villen, Michael~J MacCoss, et~al.
\newblock Carafe enables high quality in silico spectral library generation for data-independent acquisition proteomics.
\newblock {\em bioRxiv}, pages 2024--10, 2024.

\bibitem{wu2008quality}
Fang-Xiang Wu, Pierre Gagn{\'e}, Arnaud Droit, and Guy~G Poirier.
\newblock Quality assessment of peptide tandem mass spectra.
\newblock {\em BMC Bioinformatics}, 9(6):1--9, 2008.

\bibitem{wu2014novel}
Sz-Wei Wu, Tsung-Hsien Pu, Rosa Viner, and Kay-Hooi Khoo.
\newblock Novel {LC}-{MS2} {Product} {Dependent} {Parallel} {Data} {Acquisition} {Function} and {Data} {Analysis} {Workflow} for {Sequencing} and {Identification} of {Intact} {Glycopeptides}.
\newblock {\em Analytical Chemistry}, 86(11):5478--5486, June 2014.
\newblock Publisher: American Chemical Society.

\bibitem{yilmaz2024sequence}
M.~Yilmaz, W.~E. Fondrie, W.~Bittremieux, R.~Nelson, V.~Ananth, S.~Oh, and W.~S. Noble.
\newblock Sequence-to-sequence translation from mass spectra to peptides with a transformer model.
\newblock {\em Nature Communications}, 15(1):6427, 2024.

\bibitem{yilmaz2022denovo}
M.~Yilmaz, W.~E. Fondrie, W.~Bittremieux, S.~Oh, and W.~S. Noble.
\newblock \textit{De novo} mass spectrometry peptide sequencing with a transformer model.
\newblock In {\em Proceedings of the International Conference on Machine Learning}, pages 25514--25522, 2022.

\bibitem{yu2025msfragger}
Fengchao Yu, Yamei Deng, and Alexey~I. Nesvizhskii.
\newblock {MSFragger}-{DDA}+ enhances peptide identification sensitivity with full isolation window search.
\newblock {\em Nature Communications}, 16(1):3329, April 2025.
\newblock Publisher: Nature Publishing Group.

\bibitem{zhao2011combining}
Peng Zhao, Rosa Viner, Chin~Fen Teo, Geert-Jan Boons, David Horn, and Lance Wells.
\newblock Combining {High}-{Energy} {C}-{Trap} {Dissociation} and {Electron} {Transfer} {Dissociation} for {Protein} {O}-{GlcNAc} {Modification} {Site} {Assignment}.
\newblock {\em Journal of Proteome Research}, 10(9):4088--4104, September 2011.
\newblock Publisher: American Chemical Society.

\end{thebibliography}
\clearpage

\appendix

\setcounter{figure}{0}
\setcounter{table}{0}
\setcounter{section}{0}
\renewcommand{\thefigure}{S\arabic{figure}}
\renewcommand{\thesection}{S\arabic{section}}
\renewcommand{\thetable}{S\arabic{table}}

\section{Supplementary Figures}

\begin{figure*}[h!]
  \centering
    \includegraphics[height=3in]{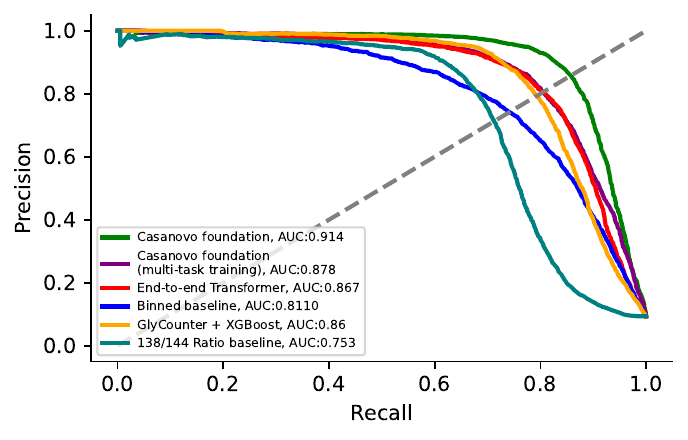}
  \caption{{\bf Glyco precision-recall curve}. Precision-recall curve for each model on the glycosylation status prediction task.
  }
  \label{fig:glyco_pr}
\end{figure*}

\begin{figure*}[h!]
  \centering
  \begin{tabular}{lll}
    \raisebox{-1\height}{\makebox[0pt][l]{\textbf{(A)}}}%
    \includegraphics[height=1.6in]{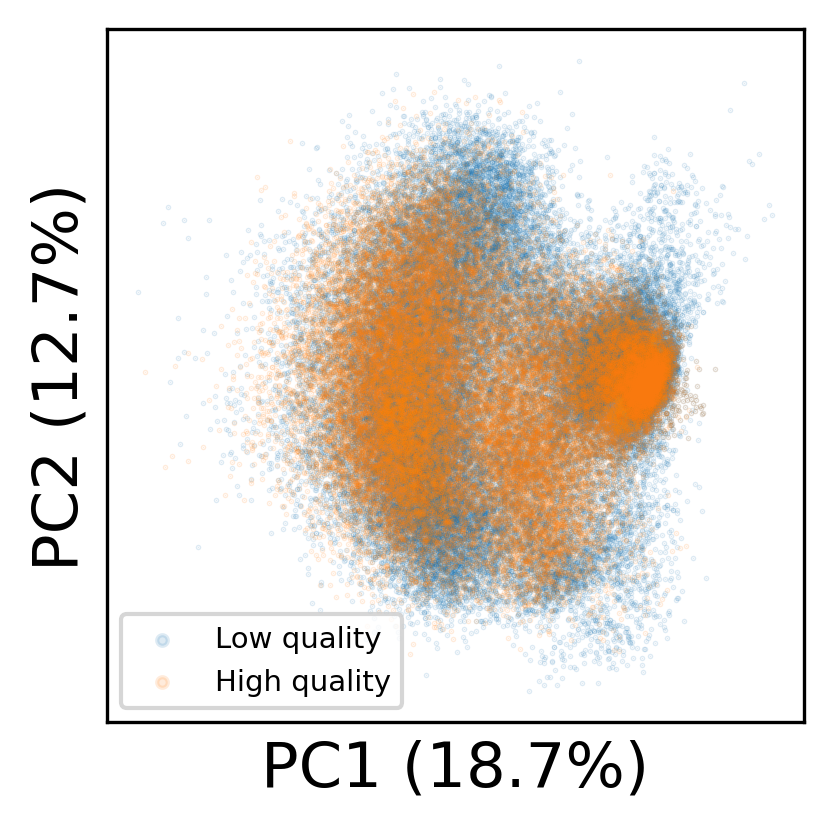} &
    \raisebox{-1\height}{\makebox[0pt][l]{\textbf{(B)}}}%
    \includegraphics[height=1.6in]{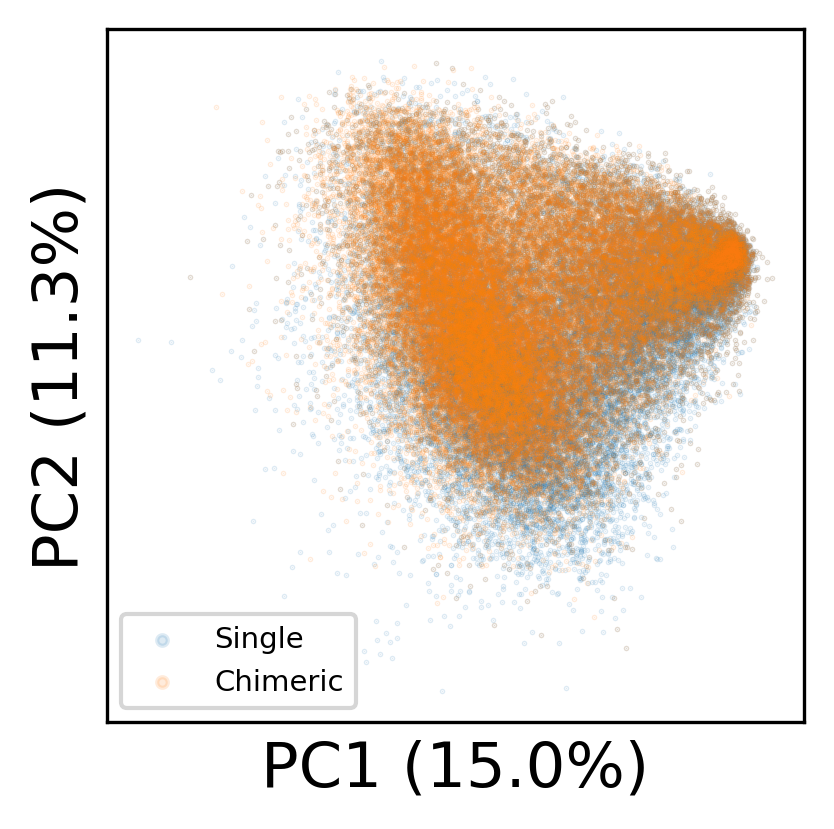} \\
    \raisebox{-1\height}{\makebox[0pt][l]{\textbf{(C)}}}%
    \includegraphics[height=1.6in]{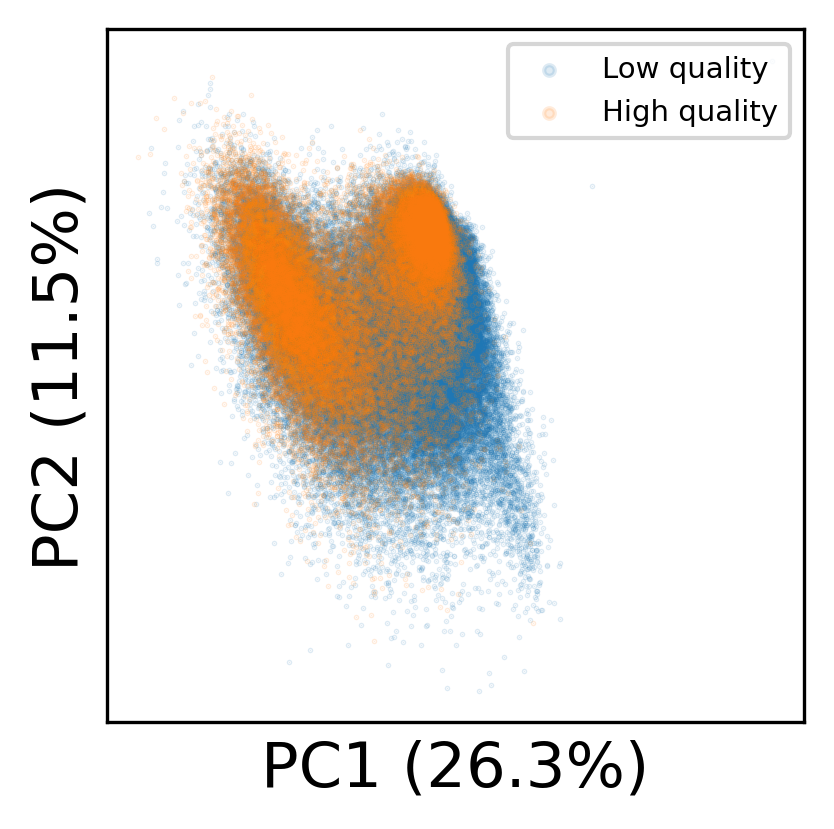} &
    \raisebox{-1\height}{\makebox[0pt][l]{\textbf{(D)}}}%
    \includegraphics[height=1.6in]{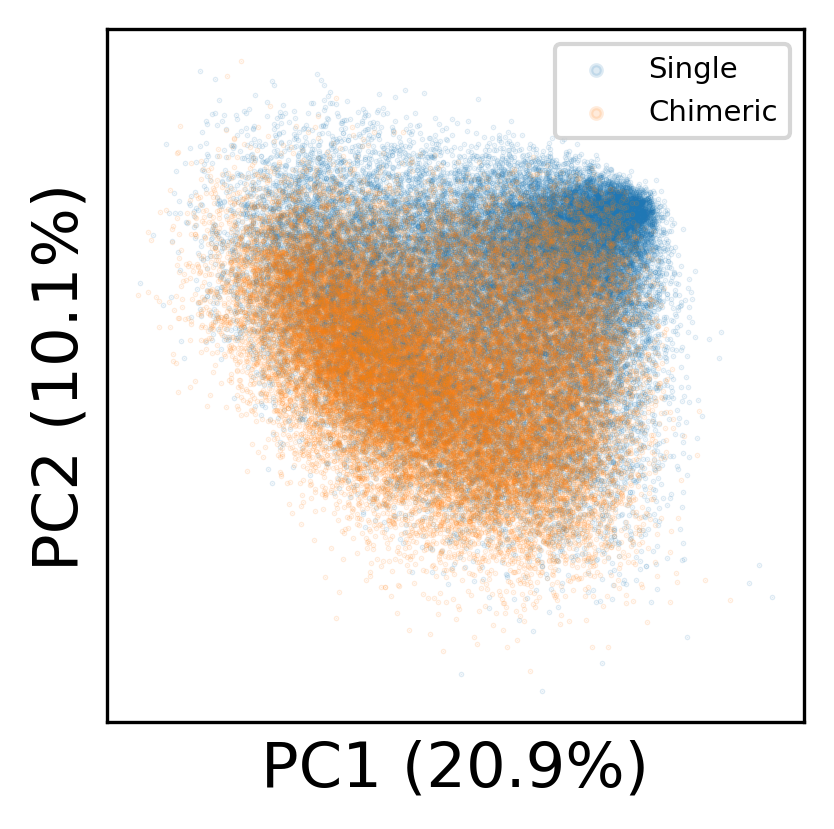} \\
  \end{tabular}
  \caption{{\bf Learned spectrum embeddings}. 
    PCA plots visualizing the learned embeddings for test set spectra. The first row contains embeddings from the pre-trained encoder for spectra from (A) the quality prediction task and (B) the chimericity prediction task. The second row (C and D) shows embeddings from the multi-task fine-tuned encoder for the same two datasets. The percentage of variance explained by each component is indicated in parentheses.}
  \label{fig:embeddings}
\end{figure*}

\begin{figure*}[h!]
  \centering
  \begin{tabular}{ll}
    \raisebox{-1\height}{\makebox[0pt][l]{\textbf{(A)}}}%
    \includegraphics[height=2.06in]{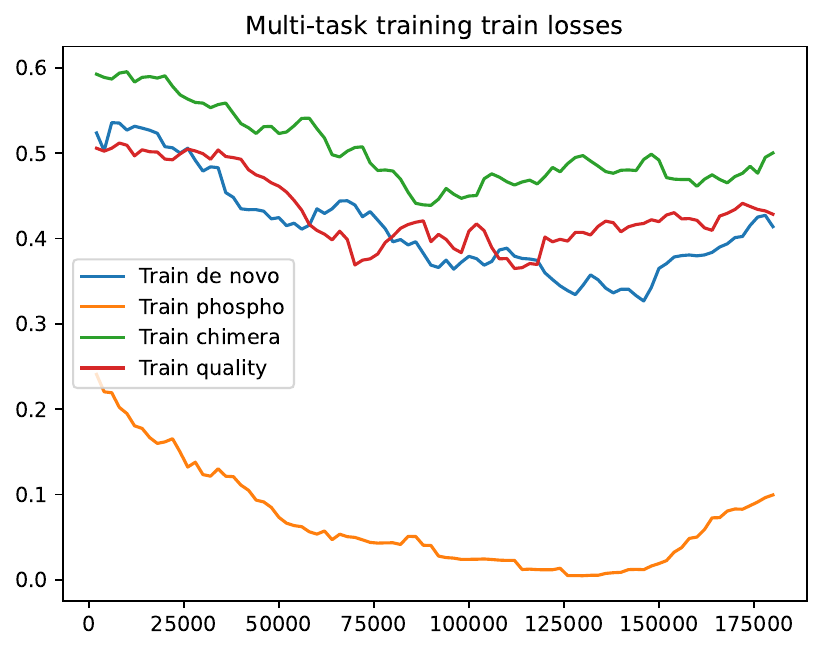} &
    \raisebox{-1\height}{\makebox[0pt][l]{\textbf{(B)}}}%
    \includegraphics[height=2.06in]{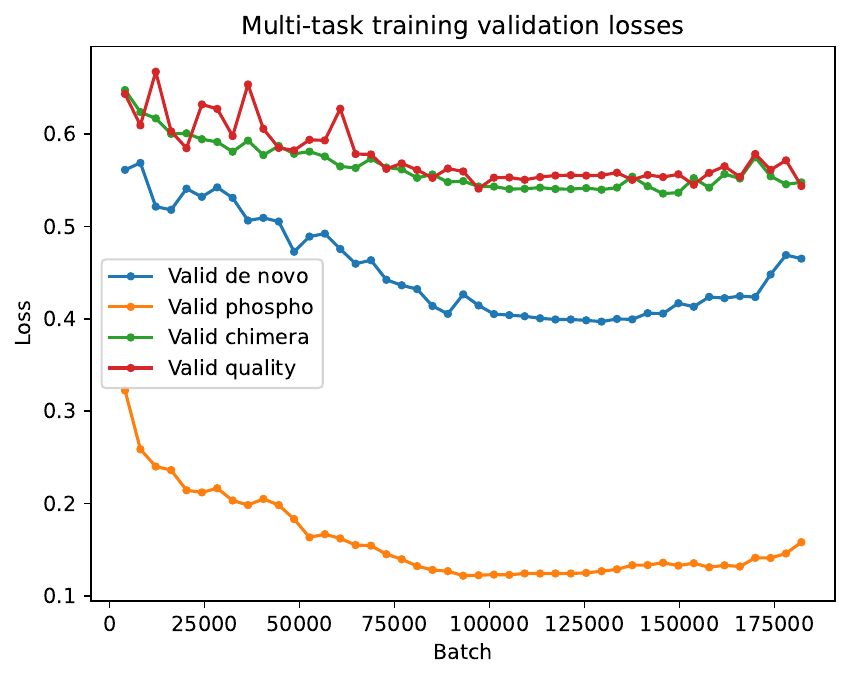} \\
  \end{tabular}
  \caption{{\bf Multi-task loss curves}. (A) Training and (B) validation loss curves for each of the four tasks during the multi-task fine-tuning of the spectrum encoder. Train losses curves are smoothed by averaging over the last 10,000 training steps.}
  \label{fig:loss_curves}
\end{figure*}

\begin{figure*}[h!]
\centering
\includegraphics[width=3in]{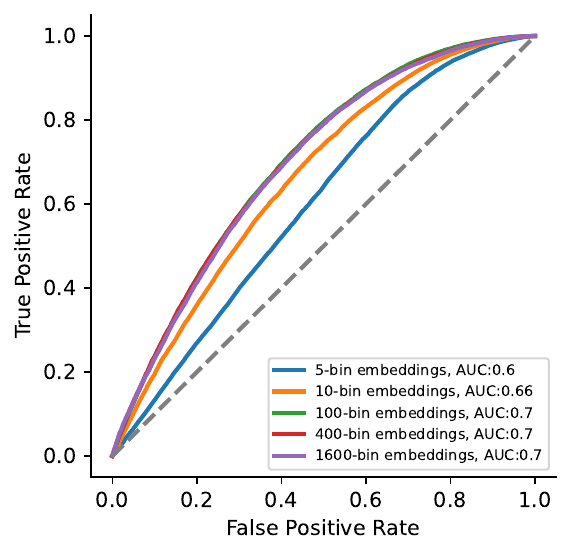}
\caption{\textbf{Comparison of binned embedding performance at different binning resolutions on the chimericity prediction.} ROC curves and the area under the curve (AUC) are reported as a function of the number of bins used to represent peak intensities for the chimericity prediction validation set.
}
\label{fig:bin_selection}
\end{figure*}

\newpage 
\section{Additional Explanations and Experimental Details}

\subsection{Quality task}
To create a labeled dataset for this task, we randomly sample 20 human Orbitrap HCD mass spectrometry runs from the MassIVE-KB data. The experiments MSV000080254, MSV000080255, MSV000081563, MSV000081607, MSV000081649, MSV000083508, MSV000083961, MSV000083966, MSV000083967 were used in the train set,   MSV000083978, MSV000083983, MSV000086369, MSV000086385, MSV000086389 in the validation set, and MSV000086439, MSV000086448, MSV000086491, MSV000088236, MSV000088405 in the test set. 
We perform a database search for each run against the reference human proteome (UniProt ID UP000005640) using Sage (version 0.14.7) with the default workflow. 
Spectra that are matched to a peptide under 1\% false discovery rate (FDR) are labeled as high quality, whereas spectra that failed to be matched are annotated as low quality.

\subsection{Chimericity task}
The samples were prepared using the method described in \cite{wen2024carafe} and analyzed using an Orbitrap Fusion Lumos mass spectrometer.
Raw MS/MS data were converted to mzML files using MSConvert with peak picking enabled in ProteoWizard (version 3.0.24031) \cite{chambers2012cross-platform}.
The human, mouse, and yeast MS/MS data were then searched against a human (20,597 proteins, 02/2024), mouse (21,701 proteins, 02/2024), and yeast (6060 proteins, 02/2024) proteome database, respectively, using FragPipe (version 22.0) with the default workflow and ``DDA+'' mode (i.e., wide window database search).
Database search results were filtered at a 1\% PSM-level FDR. Spectra were assigned as chimeric if involved in more than one high-confidence PSM. 

\subsection{Phosphorylation task}
The raw data and database search results from the human phosphoproteome dataset were downloaded from ProteomeXchange PXD012174 \cite{ochoa2020functional}, which contains data from 101 human cell and tissue types analyzed using phospho-enrichment assays. Data was prepared following the pre-processing scripts used by AHLF \cite{altenburg2022ad}, which were shared by the authors. These filtered spectra at a 1\% FDR at the PSM, protein, and phosphosite localization level. Additionally, PSMs were filtered  based on a minimum score for modified peptides of 40, and a minimum delta score for modified peptides of 6. Spectra assigned only to phosporylated peptides were assigned a positive label and spectra assigned only to unphosphorylated peptides were assigned a negative label. Remaining spectra were discarded. Finally, the data was split into train/validation/test sets at the cell/tissue type level following the same splits used by Altenberg \emph{et al} \cite{altenburg2022ad}.

\subsection{Glycosylation task}

The raw data from the 48 mouse brain HCD runs generated by Potel \emph{et al.}\ was downloaded from ProteomeXchange PXD052447 \cite{potel2025uncovering}, along with the FragPipe \cite{polasky2023msfragger} N- and O-glycopeptide search results. 
Raw MS/MS data were converted to mzML files using MSConvert with peak picking enabled in ProteoWizard (version 3.0.24031) \cite{chambers2012cross-platform}.
The data were randomly split at the run level into train/validation/test sets containing 36/6/6 runs each. 
N-glyco PSMs in the MSFragger search results were filtered for assigned modifications at asparagine residues; O-glyco PSMs were filtered for assigned modifications at either serine or threonine residues. 
From there, results were filtered based on a hyperscore greater than 16 and a glycan q-value less than 0.01 to obtain a 1\% FDR for glycan assignment. 
For confident classification labels, cases of co-occupancy of O- and N-glycosites on the same PSM were filtered out. 
Spectra identified as containing an O-glycopeptide were labeled as positive examples, while spectra containing an N-glycopeptide were assigned negative labels. 
Spectra not identified with a glycopeptide were discarded. 
GlyCounter \cite{kothlow2025extracting} was run on each of these spectra, yielding a list of 54 oxonium ions, which were in turn used to calculate the $m/z$ 138/144 ratio. 

\subsection{N- vs O-glycosylation}
\label{sec:glyco_note}
In some cases, distinguishing N-glycosylation from O-glycosylation is straightforward using ratios of ions that indicate the presence of N-acetylglycosamine (GlcNAc) or N-acetylgalactoseamine (GalNAc). This classification can be simplified to a comparison of m/z 138 to m/z 144, where a ~1:1 ratio indicates the presence of GalNAc, but not GlcNAc, in a glycan composition. This ratio is useful for classifying N-glycopeptides, which have GlcNAc but not GalNAc residues, relative to simple core 1 O-glycopeptides, which only contain GalNAc. This task becomes more challenging when considering elongated core-1 O-glycans and core 2-8 O-glycans that contain both GalNAc and GlcNac moieties. For example, core 2 glycans are relatively common in mammalian glycoproteomic datasets, and the GlcNAc residues in these O-glycopeptides mean they produce oxonium ion patterns that look more similar to N-glycopeptides than core 1 O-glycopeptides that lack GlcNAc.

\section{Training Settings and Hyper-parameters}

\subsection{Supervised pre-training}

The weights for the pre-trained Casanovo model checkpoint 4.0.0 (Apache 2.0 license) were downloaded from GitHub. This model was trained on the MassIVE-KB dataset using the supervised \textit{de novo} sequencing task as described in Yilmaz \emph{et al.}\ \cite{yilmaz2024sequence}. Only the weights for the spectrum encoder from the encoder-decoder Casanovo model were used. This gives an encoder-only model with nine transformer encoder block layers, an embedding size of 512, and eight attention heads. Overall spectrum representations were obtained from this encoder by taking the mean of the individual peak embeddings. 

\subsection{Multi-task training}
For our multi-task training experiment, we fine-tune the pre-trained spectrum encoder on three downstream tasks simultaneously. One task-specific classification head, consisting of a dense network with one 512-dimensional hidden layer and ReLu non-linearity, is added to the model for each task. Each task-specific head takes as input the mean of the individual peak embeddings output by the encoder. To prevent fine-tuned spectrum representations from overfitting to the three specific tasks, we also retain the standard \textit{de novo} sequencing loss during fine-tuning. 

When training the multi-task model, during each training step we load one batch of spectra from each task. The task-specific loss is computed on each batch respectively, and the sum of all four losses is optimized. To roughly balance the size of the datasets for the three downstream tasks, we downsampled the phosphoproteomics dataset by a factor of 32 to obtain a dataset of roughly 240,000 spectra. We anticipate that more specific re-weighting of the task-specific losses to account for differences in difficulty of each task may improve overall results by preventing one task from overfitting before other tasks have converged. However, due to resource constraints, we were unable to extensively search the space of loss weighting terms. 

We trained the multi-task model for 185,000 training steps with a batch size of 32, performing validation every 4000 steps. 
The model checkpoint with the lowest average validation loss across the three downstream tasks was selected (step 96,000). 
The peak learning rate was set to 1e-5, with a linear warmup period of 1000 steps and cosine learning rate schedule with a half-period of 120,000 steps. 
Binary cross entropy loss was used for all four tasks, with label smoothing of 0.001. Gradient updates were performed using the Adam optimizer \cite{kingma2015adam} with 1e-6 weight decay. 

\subsection{Task-specific training}

\paragraph{Binned baseline.}
To pre-process the input for our binned baseline models, we discretize the \textit{m/z} axis into equal-width bins between 150 and 2000 \textit{m/z}. For the binned embeddings, we experimented with different binning resolutions to obtain the spectrum embeddings and settled on using 100-bin, i.e.\ 100-dimensional, embeddings (Supplementary Figure~\ref{fig:bin_selection}). Peaks outside the range 140--2000 \textit{m/z} are filtered out, and the remaining peak intensities are binned at 18.6 \textit{m/z} resolution. We then train a gradient-boosted decision tree classifier on these representations \cite{chen:xgboost} using the validation set for early stopping based on validation AUROC. The hyperparameter early\_stopping\_rounds was set to 32, and 
n\_iters was chosen to be sufficiently large that training is always terminated by early stopping. Otherwise, default parameters were used. 

\paragraph{Glycounter baseline.}
Similar to the binned baseline, the GlyCounter baseline for the glycosylation status prediction task represents each spectrum as a 54-dimensional vector of intensities for a pre-defined set of oxonium ions known to be produced by glycan fragmentation. An XGBoost classifier is likewise trained on these representations. 

\paragraph{End-to-end transformer.}
For the end-to-end transformer pipeline, we train the transformer spectrum encoder and MLP classifier head end-to-end on each task. The transformer encoder is implemented using \href{https://github.com/wfondrie/depthcharge}{depthcharge} components to have the same architecture as the Casanovo encoder, except for the number of transformer layers, which was optimized based on validation set performance from the interval [1-9]. Gradient updates during were performed using the Adam optimizer \cite{kingma2015adam} with a learning rate of 1e-4 and a weight decay of 1e-6. Training is terminated with early stopping based on AUROC on the validation set with a patience of to 5 epochs. 

\paragraph{Casanovo Foundation.}
To apply Casanovo Foundation to a downstream task, we first use the pretrained encoder from Casanovo version 4.0.0 (Apache License 2.0) to obtain 512-dimensional spectrum embeddings for each spectrum. We then train a small two layer dense network on these embeddings. The dense model has one 512-dimensional hidden layer with ReLU activation. The model is trained with a learning rate of 1e-3, and is also terminated via early stopping with a patience of 5 epochs. 

\subsection{Timing and compute resources}
Pre-training Casanovo on MassIVE-KB took 8 days on 4 RTX 2080 Ti GPUs. Multi-task fine-tuning took 4 days on a single A100 80G GPU. 
Training models for each of the downstream tasks was done using 2 L40S 48GB GPUs and took $\sim$8 days in total. The majority of this time was spent training the end-to-end transformer model on the phosphorylation task. All remaining experiments were done on a CPU workstation with 16x Intel Xeon CPU E5-2680 @ 2.70GHz and 64GB of RAM in relatively negligible time. 

This dependence on large compute resources, GPUs in particular, is a notable current limitation of Casanovo Foundation. 
In practice, many mass spectrometry proteomics labs do not have access to or familiarity with using GPUs. 
However, as deep learning is becoming more widespread in the field, labs are beginning to invest more in local and cloud-based compute resources. 
Additionally, future engineering efforts to accelerate the inference time of Casanovo Foundation can further bridge this gap. 

\end{document}